\documentclass[conference]{IEEEtran}
\IEEEoverridecommandlockouts
\usepackage{cite}
\usepackage{amsmath,amssymb,amsfonts}
\usepackage{algorithmic}
\usepackage{graphicx}
\usepackage{textcomp}
\usepackage{xcolor}
\usepackage[linesnumbered,ruled,vlined]{algorithm2e}
\usepackage{booktabs}
\usepackage{multirow}
\ifCLASSOPTIONcompsoc 
	\usepackage[caption=false,font=normalsize,labelfon
t=sf,textfont=sf]{subfig}
\else
	\usepackage[caption=false,font=footnotesize]{subfig}
\fi

\def\BibTeX{{\rm B\kern-.05em{\sc i\kern-.025em b}\kern-.08em
    T\kern-.1667em\lower.7ex\hbox{E}\kern-.125emX}}
\begin{document}

\title{Swap-based Deep Reinforcement Learning for Facility Location Problems in Networks \\
\thanks{* corresponding authors: Yanyan Xu (yanyanxu@sjtu.edu.cn), Yaohui Jin (jinyh@sjtu.edu.cn).\\
This work was jointly supported by the National Key Research and Development Program (2022YFC3303102), the National Natural Science Foundation of China (62102258), the Shanghai Municipal Science and Technology Major Project (2021SHZDZX0102), and the Fundamental Research Funds for the Central Universities.}
}

\author{\IEEEauthorblockN{Wenxuan Guo}
\IEEEauthorblockA{
\textit{AI Institute}\\
\textit{Shanghai Jiao Tong University}\\
Shanghai, China \\
0000-0001-6336-3819}
\and
\IEEEauthorblockN{Yanyan Xu *}
\IEEEauthorblockA{\textit{AI Institute} \\
\textit{Shanghai Jiao Tong University}\\
Shanghai, China \\
0000-0001-5429-3177}
\and
\IEEEauthorblockN{Yaohui Jin *}
\IEEEauthorblockA{\textit{AI Institute} \\
\textit{Shanghai Jiao Tong University}\\
Shanghai, China \\
0000-0001-6158-6277}}

\maketitle

\begin{abstract}
Facility location problems on graphs are ubiquitous in real world and hold significant importance, yet their resolution is often impeded by NP-hardness. Recently, machine learning methods have been proposed to tackle such classical problems, but they are limited to the myopic constructive pattern and only consider the problems in Euclidean space. 
To overcome these limitations, we propose a general swap-based framework that addresses the p-median problem and the facility relocation problem on graphs and a novel reinforcement learning model demonstrating a keen awareness of complex graph structures. Striking a harmonious balance between solution quality and running time, our method surpasses handcrafted heuristics on intricate graph datasets.
Additionally, we introduce a graph generation process to simulate real-world urban road networks with demand, facilitating the construction of large datasets for the classic problem.
For the initialization of the locations of facilities, we introduce a physics-inspired strategy for the p-median problem, reaching more stable solutions than the random strategy. The proposed pipeline coupling the classic swap-based method with deep reinforcement learning marks a significant step forward in addressing the practical challenges associated with facility location on graphs.
\end{abstract}


\section{Introduction}
\begin{figure*}[tb]
	\centering
	\includegraphics[width=\textwidth]{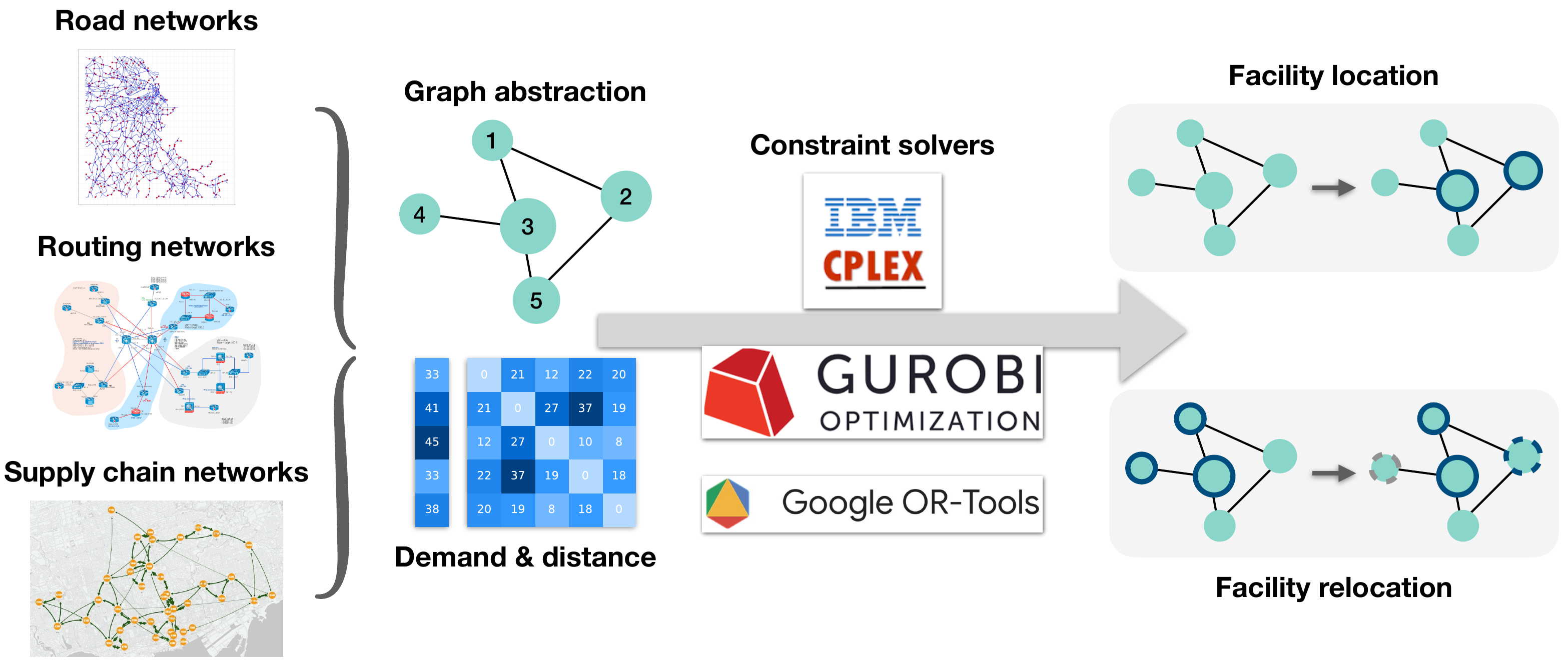}
	\caption{The general pipeline of solving real-world facility locations problems on graphs. The real-world networks are converted into abstract graph representation. The demand for each node and pairwise distance are used to encode the problems as mathematical optimization models. This quantitative representation, along with the specified constraints of some type of problem, are fed into constraint solvers, yielding the final solutions. }
	\label{fig:task}
\end{figure*}

Facility location problems study optimizing the placement of a set of facilities to satisfy the demands from customers and minimize a certain objective function. In practice, different models are used for specific needs, including single/multiple facility problems, median location problem, dynamic location problem, etc~\cite{celikturkogluComparativeSurvey2020}. 
According to~\cite{farahaniFacilityLocation2009}, there are four components that describe location problems: customers, facilities to be located, a space in which customers and facilities are located, and a distance metric between customers and facilities. 
One of the distance metric is the routing distance in networks, where customers and facilities are on the nodes of a graph. To tackle this optimization problem in non-Euclidean space, an available solution is to calculate the pairwise distance matrix between all points and transform the problem into a general formulation. 
Fig.~\ref{fig:task} describes the pipeline of solving real-world facility location problems on graphs. The real networks in different fields are first converted into graph abstraction. The demand of each node and the pairwise distances between nodes with demand and potential nodes for facilities, as the key components of an instance, are used to encode the problems as mathematical optimization models. This quantitative representation, along with the specified constraints of some type of problem, are fed into constraint solvers to yield the final solutions. 

Among various concrete problems of facility location problems in networks, we focus on two of them, namely the $p$-median problem (PMP) and the facility relocation problem (FRP). 
PMP is a crucial branch of facility location problems that seeks to minimize the weighted sum of distance costs between facilities and demand points, with fixed costs for opening facilities and a predefined number of facilities $p$. Different fields witness the applications of PMP, including designing electric charging stations networks~\cite{GAVRANOVIC201415}, establishment of public services such as schools~\cite{ndiayeApplicationPMedian2012}, and siting of shared bicycles~\cite{CINTRANO2020113684}.
Not only is PMP crucial for solving real-world problems, but it also captures the interest of researchers as a classical combinatorial optimization problem known for its NP-hardness.

While the $p$-median model is effective for static scenarios, it faces limitations in more dynamic and constrained situations. Consider urban infrastructure, like evacuation shelters, constructed in a city's early stages. The dynamic and evolving population within this area over time creates a disparity between real demands and the outdated facility layout.
Under such circumstances, it becomes more sensible to relocate facilities rather than plan anew, especially given economic constraints that limit the number of relocations. This application introduces a second problem of interest: enhancing the existing facility layout within a restricted number of relocation steps, namely the facility relocation problem.

Many heuristics and meta-heuristics have been devised to solve the facility location problems, which can be broadly categorized into two genres: constructive methods and improving heuristics. Constructive methods start from an empty set of facilities and incrementally build the solution step-by-step, while improving heuristics make modifications to feasible solutions. 
The recent advancement in machine learning, particularly deep learning, offers an alternative perspective on solving these classical problems. The expressive and generalizable nature of neural networks makes them powerful tools for tackling intricate combinatorial optimization problems~\cite{bengioMachineLearning2020}.

Previous works utilizing machine learning techniques for facility location problems, such as~\cite{wangSolvingUncapacitated2022},\cite{matisReinforcementLearning2023}, and\cite{luoFacilityRelocation2023}, predominantly follow the constructive approach of creating solutions. Empirically, the improving heuristics have better performance at the sacrifice of longer running time.
Moreover, these works adopt a simplified geometrical setting, where demands and facilities are distributed in the Euclidean space, and the Euclidean distance is used. 
In many scenarios, such as urban planning and network routing, graph distances can portray the actual traveling cost more accurately than straight-line distance. For example, the (dis)connectivity relationship and topology of urban road networks has a significant impact on urban mobility~\cite{tsiotasTopologyUrban2017}. 

To overcome these limitations, we present a comprehensive swap-based framework designed to address both the $p$-median problem and the facility relocation problem on graphs. This framework is then instantiated with two novel variants, both of which are aware of the dynamic attributes of the instances. The novel reinforcement learning model demonstrates a keen awareness of complex graph structures of the original instances, and thus makes improvement decisions effectively and efficiently. 
In addition to the swap-based framework, we introduce a physics-inspired initialization strategy for the $p$-median problem based on the scaling law, which implies a general relation between distributions of demands and facilities in large networks.
Furthermore, recognizing the importance of realistic graph generation, we ingeniously design a method that adheres to urban distribution patterns. This ensures that the generated graphs closely mimic real-world scenarios, contributing to the validity and relevance of our approach.

\section{Related Work}
In this section, we provide an overview of the recent advancements in applying machine learning techniques to tackle combinatorial optimization problems. The field has witnessed significant progress in developing intelligent algorithms capable of solving complex optimization challenges. We begin by surveying the general landscape of machine learning solutions for combinatorial optimization, with an emphasis on reinforcement learning for constructive algorithms and improving algorithms. Subsequently, we narrow our focus to the specific domain of facility location and relocation problems.

%

\subsection{Machine Learning for Combinatorial Optimization}
Combinatorial optimization (CO) refers to a class of optimization problems where the solutions are defined in a discrete set. Specifically, we are concerned about the CO problems with NP hardness, for there are currently no general efficient (with polynomial complexity) algorithms for solving these problems. Typical examples of such problems are the traveling salesman problem (TSP), vehicle routing problem (VRP), job shop scheduling problem (JSSP), etc. 
The survey~\cite{bengioMachineLearning2020} provides a comprehensive review on the intersection of machine learning and combinatorial optimization. From the aspect of how machine learning methods aid the optimization process, we mention two typical paradigms. 
The first line of works train end-to-end learning frameworks towards a differentiable loss function with supervision, which is the objective function of the original problem or a surrogate goal. The other pipeline of solving uses machine learning model as a ``consultant'' that provides decisions for a master algorithm. 

\paragraph{Reinforcement Learning for Solution Construction of Graph CO problems}
Among various combinatorial optimization problems, we pay extra attention to the ones that can be formulated on graphs. For one thing, the graph structures are ubiquitous in the real world and have wide applications in many fields, and many CO problems are naturally defined in the graph context, such as minimum vertex cover and graph matching. For another thing, graph neural networks are well studied and have strong abilities of representation, thus they can be a strong weapon dealing hard problems on graphs. 
Reinforcement learning has been employed to address solution construction processes in combinatorial optimization, especially in the context of graph problems. Representative works in this domain include~\cite{daiLearningCombinatorial2018}, which trains a greedy node selection policy using a Graph Neural Network (GNN) as a scorer with the DQN algorithm. Another example is~\cite{kool2019attention}, leveraging the attention mechanism to sequentially predict the next node in routing problems. A comprehensive review of works in this area is available in the survey~\cite{cappartCombinatorialOptimization2022}.

\paragraph{Machine Learning for CO Solution Improvement}
Broader in scope, some works explore general improvement-style algorithms for combinatorial optimization problems using machine learning methods. 
The pioneering work~\cite{chenLearningPerform2019} introduces NeuRewriter, a reinforcement learning model that learns region-picking and rewriting-rule policies. This model is applied to expression simplification, job scheduling, and capacitated vehicle routing problems. 
In~\cite{luLearningbasedIterative2019}, the focus is on enhancing solutions to the capacitated vehicle routing problem, incorporating perturbation operators for a larger search space. The model includes the running history as a feature for the networks.
Additionally, in~\cite{wuLearningImprovement2022}, improving heuristics for two routing problems are considered. Node pairs in the action space are scored by a compatibility layer computed based on query and key from self-attention layers. 
Summarizing three intervention points of meta-heuristics, \cite{falknerLearningControl2023} designs a policy model based on graph neural networks to assist local search, conducting experiments on job shop scheduling and capacitated vehicle routing problems. 
It's noteworthy that most works in this domain primarily focus on routing problems, which exhibit different features from facility location problems.

\subsection{Solving Facility Location Problems}
We focus on the $p$-median problem as a case of facility location problems. 
Given the extensive history and practical importance of the $p$-median problem, numerous heuristics and algorithms have been developed. We first give a brief introduction of the meta-heuristics and heuristics, and then review recent advance on facility location problems with the help of machine learning techniques.
 
\paragraph{Approximation Algorithms for the P-median Problem}
Many meta-heuristics have been applied to this problem, including variable neighborhood search~\cite{hansen1997variable}, genetic algorithms~\cite{alp2003efficient}, tabu search~\cite{rolland1997efficient}, etc. Since the p-median problem can also be formulated as an integer programming problem, it is therefore compatible with Lagrangian relaxation methods~\cite{cornuejols1977exceptional} and general solvers such as gurobi~\cite{gurobi}. 
Other manually designed heuristics include greedy addition~\cite{kuehn1963heuristic}, alternate selection and allocation~\cite{maranzana1964location}, and exchange algorithm~\cite{goodchild1983location}. Readers are referred to~\cite{reeseSolutionMethods2006} for a comprehensive survey of solution methods for the p-median problem.

\paragraph{Machine learning for Facility Location Problems}
Although the facility location problems have a long history and many heuristics have been developed, there are relatively few works that approach this classical problem with machine learning, especially deep learning methods. Most current works are constructive and deploy the reinforcement learning techniques.
\cite{wangSolvingUncapacitated2022} first proposes to solve the uncapacitated $p$-median problem in the Euclidean space with reinforcement learning and graph attention networks. It falls into the category of constructive algorithms, and it formulates the solution construction scheme as a Markov decision process using the REINFORCE~\cite{williams1992simple} algorithm to choose the next facility in the solution. The core neural module in~\cite{wangSolvingUncapacitated2022} is graph attention networks with multi-talking-heads, and it predicts the next point in solution with an encoder-decoder architecture. 
Another work that solves the weighted $p$-median problem with reinforcement learning is~\cite{matisReinforcementLearning2023}. It has a similar formulation as~\cite{wangSolvingUncapacitated2022} that builds a solution constructively. It tests with different implementation of convolutional neural networks. However, the models in~\cite{matisReinforcementLearning2023} were trained and tested on the same dataset, and generalizability is not discussed. Besides the $p$-median problem, there are works on other branches of the facility location problems, such as~\cite{shaohuawangDeepMCLPSolving2023} that solves the maximal covering location problem constructively with reinforcement learning. 
As for the relocation problem, \cite{luoFacilityRelocation2023} addresses the facility relocation problem with a twofold objective of facility exposure and user convenience. The authors uses a reinforcement learning module as an assistive component to a greedy algorithm that maximizes the single-step reward. The action space for the agent is very small, as it only predicts a binary option of whether to relocate a facility.

Our work is the first deep reinforcement learning method that addresses the facility location problem from an improving perspective and jointly solves the relocation problem. Our agent has a much higher level of autonomy compared to~\cite{luoFacilityRelocation2023} of choosing which facility to relocate as well as the destination, and demonstrates robust generalizability in the meanwhile.


\section{Preliminaries and Problem Formulation}
\label{sec:preliminaries}

\begin{table}[tb]
  \centering
  \caption{Frequently Used Symbols and Notations}
  \label{tab:symbol}
  \begin{tabular}{|c|l|}
    \hline
    \textbf{Symbol} & \textbf{Definition} \\
    \hline
    $G(V, E)$ & A graph with nodes $V$ and edges $E$ \\
    \hline
    $n=|V|$ & The number of nodes \\
    \hline
    $F$ & Facility set \\
    \hline
    $P=\{p_{i}\}, i \in V$ & User demand \\
    \hline
    $D=\{d_{ij}\}, i, j \in V$ & Distance matrix \\
    \hline
    $p$ & The number of facilities \\
    \hline
    $F_0$ & Existing facility set for relocation \\
    \hline
    $Q$ & Improvement ratio for relocation \\
    \hline
    $V_r, r \in F$ & The Voronoi cell of the $r$-th facility \\
    \hline
    $\rho(\mathbf{r})$ & Population density at position $\mathbf{r}$ \\
    \hline
    $D(\mathbf{r})$ &  facility density at position $\mathbf{r}$ \\
    \hline
  \end{tabular}
\end{table}

In this paper, we consider the facility location and relocation problems on undirected graphs. We first introduce the formulation and heuristics of the classic $p$-median problem, and then extend the problem to facility relocation problem, along with other necessary definitions such as Voronoi cells and scaling law in urban settings.

\subsection{P-median Problem}

\subsubsection{Formulation}

Given a graph $G(V,E)$ comprising nodes $V$ and edges $E$, the objective of the $p$-median problem on $G$ is to select a facility set $F$ of size $p$ to minimize the overall cost, defined as the weighted sum of nodes to their nearest facilities. In this context, users are positioned on the nodes, and all nodes are potential facilities. Hence, the user set encompasses all nodes in $V$, and the potential facility set comprises all unoccupied nodes. In the case of the $p$-median problem, no facilities exist initially.
Each facility possesses infinite capacity, and a fixed $p$ is predetermined before problem resolution. Consequently, our focus is solely on the travel cost and user demands. For ease of expression, let $n=|V|$ denote the number of nodes. Each node $i$ has a demand of $p_i \ge 0$, collectively represented as $P = \{p_i\}$.
The travel cost between nodes is defined as the shortest paths on graphs, expressed through the distance matrix $D \in \mathbb{R^+}^{n\times n}$, where $d_{ij}$ signifies the distance between nodes $i$ and $j$. Conveniently, the distance matrix can be computed using Dijkstra's algorithm~\cite{dijkstraNoteTwo1959}.
The objective function of the $p$-median problem, denoted as $\mathcal{O}(F)$, and the optimal facility set $F^*$ are formally articulated as follows:
\begin{align}
    \mathcal{O}(F) &= \sum_{i\in V} p_i \min_{j\in F} d_{ij}, \quad \mathrm{s.t.\ } F\subseteq V, |F|=p. \label{eq:obj_pmp} \\
    F^* &= \mathop{\arg\min}_{F} \mathcal{O}(F).
\end{align}

An alternative formulation of the $p$-median problem is presented through the subsequent integer programming model:
\begin{align} 
    \min & \sum_i \sum_j d_{ij} x_{ij},\quad \mathrm{s.t.}\\ 
    &\sum_j x_{ij} = 1, \forall i, \\ 
    &x_{ij} \le y_j, \forall i,j, \\ 
    &\sum_j y_j = p, \\ 
    &x_{ij}, y_j \in \{0,1\},
\end{align}
where $y_j$ denotes whether a facility is opened in node $j \in L$, and $x_{ij}$ indicates whether user $i$ is served at facility $j$.

\subsubsection{Heuristics}
\label{sec:pmp_heuristics}
The $p$-median problem is an extension of the Weber problem, requiring the identification of a point that minimizes the total distance to all demand points. Despite a seemingly minor alteration involving the increase in the number of facilities, this adjustment significantly impacts the problem's complexity, as evidenced by its NP-hard status~\cite{kariv1979algorithmic}. In practical terms, there are no polynomial algorithms available for precise solutions. Evaluating all $\binom{n}{p}$ possible solutions becomes impractical as both $n$ and $p$ grow, leading to the adoption of heuristics and approximate algorithms.

Given the extensive history and practical importance of the $p$-median problem, numerous heuristics and algorithms have been developed for effective and efficient solving. Three classical solving schemes serve as prototypes for many heuristics in this domain. 
The first is the greedy-adding algorithm, initiating with an empty set of facilities and progressively incorporating the optimal ones. 
Another heuristic is the alternate or Maranzana~\cite{maranzana1964location} algorithm, alternating between assigning users to the nearest facilities and solving 1-median problems in the vicinity of each facility. 
The third prototype is the vertex substitution or interchange algorithm, commencing with a feasible solution and iteratively enhancing it through local exchanges by substituting some facilities with other candidates.

From a different perspective, heuristics can be broadly categorized into two genres: constructive methods and improving heuristics. Constructive methods commence with an empty set of facilities and incrementally build the solution step-by-step. Greedy-adding and alternate algorithms fall into this category. 
In contrast, improving heuristics aim to enhance a feasible solution, akin to the vertex substitution algorithm. Constructive methods typically exhibit myopic behavior by not altering or undoing chosen facilities. Empirical results corroborate this observation, indicating that interchange algorithms achieve lower optimality gaps~\cite{gwalaniEvaluationHeuristics2021}, albeit at the expense of increased runtime.

Given these considerations, we advocate for an improving algorithmic approach, utilizing machine learning techniques to identify beneficial modifications to solutions with enhanced efficiency. Notably, the interchange algorithm stands out as its solving process can also be viewed as addressing the relocation problem.
Consequently, we propose an improving algorithm employing deep reinforcement learning, addressing both the relocation problem independently and serving as a functional module for solving the $p$-median problem from scratch.


\subsection{Voronoi Cells and Scaling Law}
\label{sec:power_law}

\begin{figure}[tb]
	\centering
	\includegraphics[width=\columnwidth]{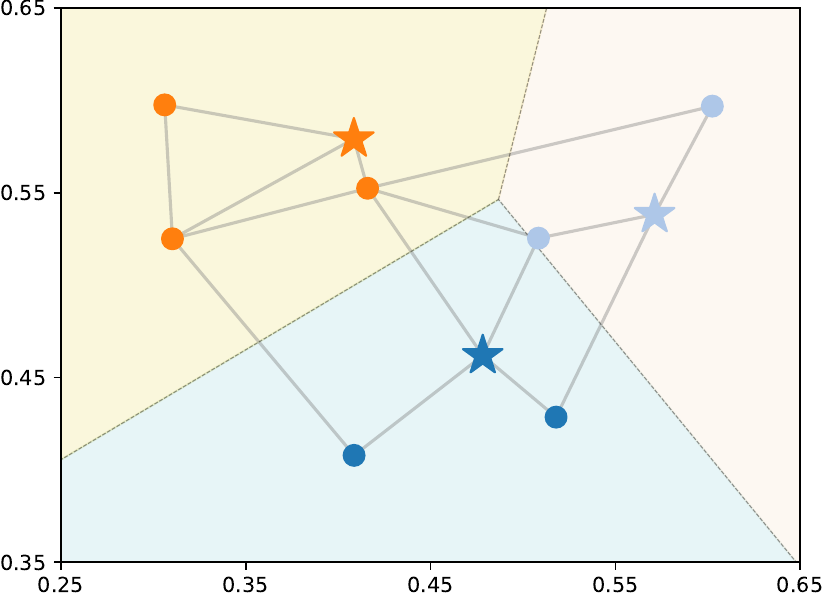}
	\caption{Demonstration of the optimal solution to a $p$-median problem instance on graph with $n=10$ and $p=3$. The stars indicate the nodes where facilities are built, and different colors of nodes represent the nearest facility to them. The plane is divided into three Voronoi cells by the facilities, distinguished by colors, which are useful for computing node features. }
	\label{fig:voronoi}
\end{figure}

In the context of facility location problems on a plane, the plane can be divided into $p$ regions corresponding to a set of $p$ facilities. Each region, termed a Voronoi cell, comprises points closest to the respective facility compared to others, as depicted in Fig.~\ref{fig:voronoi}. Voronoi cells provide insights into the quality of facility placement, and optimal assignments exhibit certain patterns under specific conditions. Notably, in the urban planning scenarios, physicists observed that there is a scaling law between population and facility densities, with exponents correlated to facility types~\cite{umScalingLaws2009}. The scaling law is briefly outlined here, with detailed proofs available in~\cite{umScalingLaws2009} and~\cite{gastnerOptimalDesign2006}.

Let $V_r$ denote the $r$-th Voronoi cell (associated with the $r$-th facility). The area of $V_r$ is denoted as $s_r$, and the demand (e.g. population in urban planning scenarios) in $V_r$ is denoted as $n_r$. The population density $\rho(\mathbf{r})$ and facility density $D(\mathbf{r})$ at position $\mathbf{r}$ are defined as $\rho(\mathbf{r})=n(\mathbf{r})/s(\mathbf{r})$ and $D(\mathbf{r})=1/s(\mathbf{r})$, where $n(\mathbf{r})=n_r$ and $s(\mathbf{r})=s_r$ if $\mathbf{r} \in V_r$. Assuming a constant geometric factor between cells and optimal facility distribution, there exists a relation between $\rho(\mathbf{r})$ and $D(\mathbf{r})$ given by
\begin{align}
	D(\mathbf{r})=\frac{1}{s(\mathbf{r})}= [\rho(\mathbf{r})]^{2 / 3} \frac{\int_A[s(\mathbf{r})]^{-1} \mathrm{~d}^2 r }{\int_A[\rho(\mathbf{r})]^{2 / 3} \mathrm{~d}^2 r}.
\end{align}
This relation can be succinctly expressed as $D \propto \rho^{2/3}$.

Building upon this observation, we introduce the \textit{VSCA} interchange algorithm for facility location problems, leveraging the total cost within each Voronoi cell, as detailed in Section~\ref{sec:swap_algs}. The scaling law also inspires us to devise a density-based initialization scheme for better optimization (see Section~\ref{sec:pmp}).

\subsection{Facility Relocation Problem}
Different from the classical $p$-median model, the facility relocation problem considers a dynamic demand changed over time and the facilities should be relocated correspondingly to meet people's needs. 
For a predefined set of facilities $F_0\subset V$ and a limited budget $k$, we study the improvement gained by moving at most $k\le|F_0|$ facilities in $F_0$. This relocation is represented by a pair of sets $(R_k, I_k)$. The updated facility set is defined as $F=F_0 \cup I_k \setminus R_k$. The fundamental assumptions regarding travel cost and demands remain consistent with the $p$-median model.
Formally, the objective of relocation problem $\mathcal{O}_k(R_k, I_k|F_0)$ can defined based on (\ref{eq:obj_pmp}):
\begin{align}
    \mathcal{O}_k(R_k, I_k|F_0) &= \mathcal{O}(F_0 \cup I_k \setminus R_k), \quad \mathrm{s.t. }  \\
    &R_k\subset F_0, \\
    &I_k \subset V\setminus F_0, \\
    &|R_k|=|I_k| \le k.
\end{align}
We can further define the improvement ratio $Q$ of a relocation set pair as the ratio of decreased cost to original cost before relocation:
\begin{align}
	Q(R_k, I_k|F_0) &= \frac{\mathcal{O}(F_0) - \mathcal{O}_k(R_k, I_k|F_0)}{\mathcal{O}(F_0)}. \label{eq:improvement}
\end{align}
In our experiments, we utilize $Q$ as a metric to evaluate solution quality, where higher values of $Q$ indicate superior relocation plans.

\section{Methods}
We start with a general swap-based framework for solving the facility relocation problem, as shown in Algorithm~\ref{alg:general_swap}. This framework is then instantiated with four variants, including two simple heuristics and our proposed methods: 
\textit{VSCA} and a deep reinforcement learning method \textit{PPO-swap}. 

\subsection{Swap-based Improvement Algorithms}
\label{sec:swap_algs}

\begin{algorithm}[tb]
	\caption{A general swap framework for relocation}
    \label{alg:general_swap}
 
    \SetKwInOut{Param}{Parameters}
    \SetKwFunction{FSwap}{SwapRelocate}
    \SetKwProg{Fn}{Function}{}{}
	\Param{iteration number $T$, swapping model $agent$}
	\KwIn{existing facilities $F_0$, relocation budget $k$}
	\KwOut{relocation pair $(R_k, I_k)$, new facility set $F$}

    \Fn {\FSwap{$T, agent, F_0, k$}} {
        $R_k \leftarrow \emptyset$\;
    	$I_k \leftarrow \emptyset$\;
    	$cost_{min} \leftarrow \infty$\;
    	
    	\For{$i \leftarrow 1$ \KwTo $T$}{
    		$R \leftarrow \emptyset$\tcp*{removed facilities} 
    		$I \leftarrow \emptyset$\tcp*{inserted facilities} 
            $F \leftarrow F_0$\tcp*{current facilities}
    		$state \leftarrow$ get\_state($F$)\;
    		
    		\For{$j \leftarrow 1$ \KwTo $k$}{
    			$(u_1, u_2) \leftarrow$ $agent$.act($state$)\;
    			
                \If{update criteria is met}{
                    $R \leftarrow R \cup \{u_1\}$\;
        			$I \leftarrow I \cup \{u_2\}$\;
        			$F \leftarrow F \setminus \{u_1\} \cup \{u_2\}$\;
        			$state \leftarrow$ get\_state($F$)\;
        			
        			\If{$\mathcal{O}(F) < cost_{min}$}{
        				$R_k, I_k \leftarrow R, I$\;
        				$cost_{min} \leftarrow \mathcal{O}(F)$\;
        			}
                }
    		}
    	}
    	\KwRet $(R_k, I_k), F$\;
    }
\end{algorithm}

\begin{table*}[tb]
    \centering
    \caption{Description of Methods and Update Criteria}
    \label{tab:swapping}
    \begin{tabular}{|c|c|c|}
        \hline
        \textbf{Method} & \textbf{Action} & \textbf{Update Criteria} \\
        \hline
        Random-swap & Random $(u_1, u_2) \in F \times  (V\setminus F)$ & \textit{True}  \\
        \hline
        Greedy-swap & $(u_1, u_2) = \mathop{\arg\min}_{(u_1, u_2)\in  F \times  (V\setminus F)} \mathcal{O}(F \setminus \{u_1\} \cup \{u_2\})$ & $\mathcal{O}(F \setminus \{u_1\} \cup \{u_2\}) < \mathcal{O}(F)$ \\
        \hline
        VSCA & $(u_1, u_2) = \mathop{\arg\min}_{(u_1, u_2)\in V_{low} \times  V_{high}} \mathcal{O}(F \setminus \{u_1\} \cup \{u_2\}) $ & $\mathcal{O}(F \setminus \{u_1\} \cup \{u_2\}) < \mathcal{O}(F)$  \\
        \hline
        PPO-swap & Sample from $P(u_1, u_2|\Theta_{\mathcal{A}})$ & \textit{True} \\
        \hline
    \end{tabular}
\end{table*}

Algorithm~\ref{alg:general_swap} gives a general swap framework for facility relocation problem. Given the set of existing facilities $F_0$ and the maximum number of relocation $k$, it incrementally builds the relocation set pair with the instructions of the given agent. 
The global solution is updated with current relocation pair when the objective function reaches a lower value.
In Algorithm~\ref{alg:general_swap}, the swapping operation is performed at most $k$ times for each run, and the iteration number $T$ is meaningful for nondeterministic algorithms, such as Random-swap. There are two major design points for different methods: the agent action logic and the stopping criteria. This framework is flexible and compatible with various algorithms, including classical handcrafted heuristics. 
We first introduce two straightforward implementations of this framework below, followed by two novel algorithms \textit{VSCA} and \textit{PPO-swap} designed by us.
A comparative analysis of the key components of these implementations is presented and summarized in Table~\ref{tab:swapping}.

\paragraph{Random-swap}
Random-swap is a naive baseline that instantiates the above swapping framework. In this method, a facility is randomly selected along with a candidate facility and this process is repeated for a total of $k$ iterations. 

\paragraph{Greedy-swap}
Greedy heuristics are simple and intuitive, and they can provide good approximated solutions in practice. The main idea is to choose the best swap at each step, i.e. the swap pair that decreases the objective function most among all possible pairs. The major drawback of this paradigm is that the overhead for evaluating each swapping pair increases too fast as the instance size grows. 


\subsection{VSCA: Voronoi-based Swap with Cost Awareness}
Given the unique correspondence between Voronoi cells and facilities, and considering that Voronoi cells augment the limited information provided with facility positions, it would be beneficial to the solving process to exploit the abundant information concealed within Voronoi cells. 
Following the definitions in Section~\ref{sec:power_law} and~\cite{umScalingLaws2009}, an important attribute of Voronoi cells is the total cost within the cell. Define $c_r$ as the total cost of $V_r$ as the summation of travel distance from every visitor to the $r$-th facility:
\begin{align}
	c_r=\sum_{u \in P\left(V_r\right)}p_u d_{f(r) u}, \label{eq:vsca}
\end{align}
where $P(V_r)$ is the set of nodes in $V_r$, and $f(r)$ is the index of the $r$-th facility.

To reach the scaling law in Section~\ref{sec:power_law}, $c_r$ is assumed to be steady across different regions, i.e., $c_r = c$. Conversely, an unbalanced distribution of $c_r$ suggests a potential relocation direction. A high cost within a region indicates the need for an additional facility in its vicinity to alleviate the high travel costs for visitors. On the contrary, a low $c_r$ implies a surplus of facilities in that region. 
Therefore, we propose an interchange strategy based on Voronoi cell costs, called \textit{Voronoi-based Swap with Cost Awareness (VSCA)}, as shown in Algorithm~\ref{alg:vsca}. Define $V_{low}$ as the Voronoi cell with lowest regional cost, and $V_{high}$ with highest cost. At a higher level, we should move the facility in $V_{low}$ to somewhere near $V_{high}$. This is also consistent with~\cite{umScalingLaws2009} from the micro-dynamics aspect. After multiple iterations of relocation, the total cost within different regions should become more balanced, typically resulting in a lower value of global objective function. 

\begin{algorithm}[tb]
	\caption{VSCA: Voronoi-based Swap with Cost Awareness}
	\label{alg:vsca}
	\KwIn{Current facilities $F$}
	\KwOut{Relocation pair $(u_1, u_2)$}
	
	\For{$r$ in range($|F|$)}{
	
		Compute the Voronoi cell $V_r$ of $r$-th facility\;
	    Compute regional cost $c_r$ of $V_r$ using (\ref{eq:vsca})\;
	}
	
	$V_{\text{high}} \leftarrow$ Voronoi cell with highest $c_r$\;
	$V_{\text{low}} \leftarrow$ Voronoi cell with lowest $c_r$\;
	
	$u_1 \leftarrow$ the facility of $V_{\text{low}}$\;
	$u_2 \leftarrow$ None\;
	$cost_{min} \leftarrow \infty$\;
	
	\For{$j$ in $P(V_{\text{high}})$}{
	    \If{$\mathcal{O}(F\setminus\{u_1\} \cup \{j\}) < cost_{min}$}{
	        $cost_{min} \leftarrow \mathcal{O}(F\setminus\{u_1\} \cup \{j\})$\;
	        $u_2 \leftarrow j$\;
	    }
	}
	
	\Return $(u_1, u_2)$\;
\end{algorithm}

\subsection{PPO-swap: a Learning-based Interchange Algorithm}
\begin{figure*}[tb]
	\centering
	\includegraphics[width=.9\textwidth]{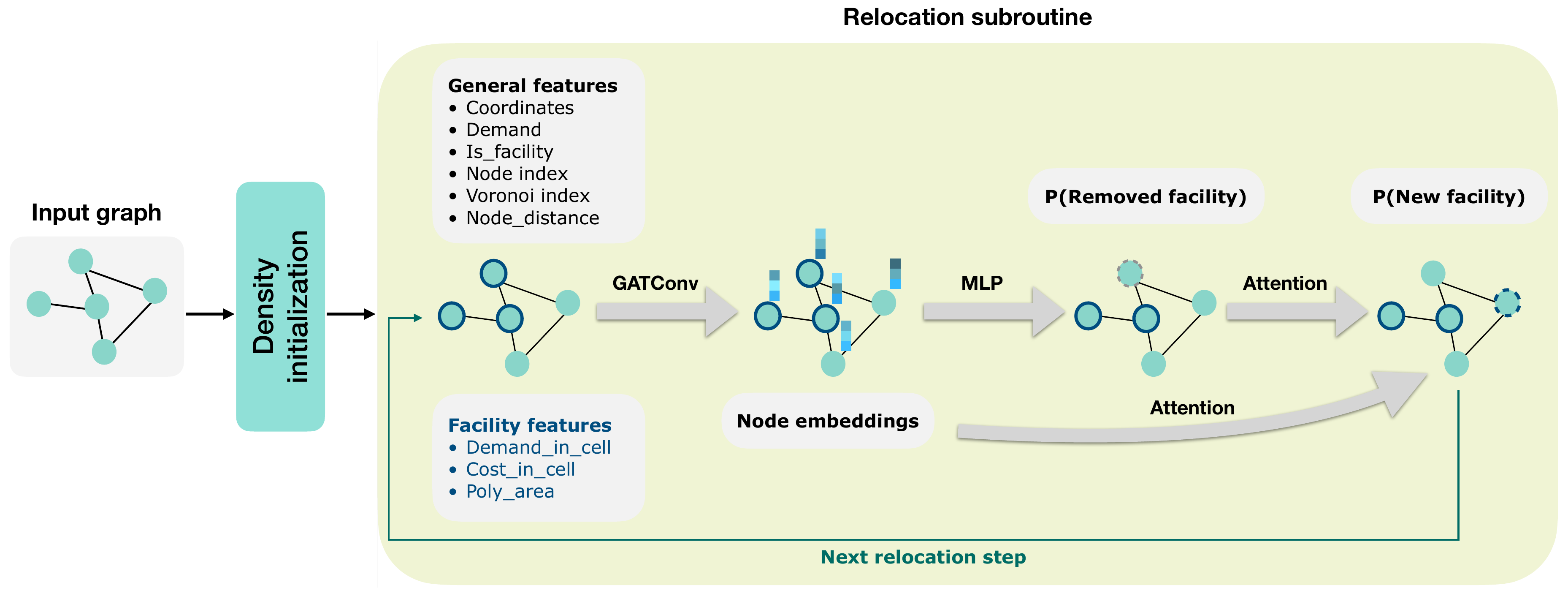}
	\caption{The pipeline of reinforcement learning based method for facility location problems. For the $p$-median problem that requires to choose $p$ facilities from scratch, we construct a feasible solution with density initialization, and iteratively optimize it with the relocation subroutine. For the relocation problem, the GAT module encode node features and yield node embeddings, which are subsequently used for computing the probability of removing and inserting facilities at each node. }
	\label{fig:ppo_pipeline}
\end{figure*}

To address the limitations of the greedy algorithm and other heuristics, we introduce a novel deep reinforcement learning method named \textit{PPO-swap}. This approach predicts the nodes to be swapped at each step using trained neural networks. In contrast to Greedy-swap and VSCA, PPO-swap permits the objective function to increase during swapping with the aim of achieving long-term benefits. The workflow of PPO-swap is elaborated in the following section. Fig.~\ref{fig:ppo_pipeline} illustrates how PPO-swap solves the facility location problems.

\subsubsection{Reinforcement Learning Formulation for Facility Relocation}

We model the relocation pair selection as a Markov decision process, with key components of reinforcement learning defined as follows. The states encompass static attributes of the instances, such as node coordinates and demand, along with the dynamic statistics associated with the current feasible solution (refer to Section~\ref{sec:gat} for details). The action space is $F \times (V\setminus F)$. Let $(u^i_1, u^i_2)$ be the relocation pair at the $i$-th step. The reward $r_i$ for this round is defined as the improvement ratio gained by executing this relocation, namely
\begin{align}
	r_i = Q(R\cup\{u^i_1\}, I\cup\{u^i_2\}|F_0) - Q(R, I|F_0).
\end{align}

For reinforcement learning training, we employ the proximal policy optimization (PPO) algorithm~\cite{schulman2017proximal}, which features a controllable update strategy by penalizing significant deviations between consecutive policies. 
Our PPO implementation adopts an actor-critic network architecture.
The actor consists of $L_1$-layer graph attention networks (GAT) for node embedding, $L_2$-layer multi-layer perceptrons (MLP) for scoring, and an extra attention layer to choose node pair conditionally. The critic has a similar structure except for the last attention layer, without weight sharing as suggested by~\cite{huang202237}. 

For convenience, we define the learnable parameters of the actor as $\Theta_{\mathcal{A}} = \Theta_{GAT} \cup \Theta_{MLP} \cup \Theta_{Att}$, and the parameters of the critic as $\Omega_{\mathcal{C}} = \Omega_{GAT} \cup \Omega_{MLP}$. 
Let $\mathbf{h}_i^l$ be the node embedding after $l$ layers of GAT for node $i$, $\mathbf{g}_i^l$ for MLP, and $\mathbf{f}_i$ for the attention layer. Specially, $\mathbf{h}_0^l$ refers to the initial features fed into the GAT. The high-level representation of the actor logic is as follows:
\begin{align}
	\mathbf{h}_i^l &= \mathbf{GAT}(\mathbf{h}_i^{l-1}), \\ 
	\mathbf{g}_i^0 &= \mathbf{MLP}(\mathbf{h}_i^{L_1}), \\
	\mathbf{g}_i^l &= \mathbf{MLP}(\mathbf{g}_i^{l-1}), \\
	\mathbf{f}_i &= \mathbf{Att}(\mathbf{h}_i^{L_2}),
\end{align}
and the operations of each module are elaborated in the following section. 

\subsubsection{Node Features and Embedding}
\label{sec:gat}
This section starts with node feature extraction, followed by the implementation of graph attention networks, namely the $\mathbf{GAT}$ module. 

For each node $i$, the raw input to GAT is a node feature vector $\mathbf{h}_i^0 \in \mathbb{R}^{10}$, which is the concatenation of general features and facility features. 
The first three features are basic attributes of nodes and remain unchanged during the solving process, including node coordinates $(x_i, y_i)$ and node demand $p_i$. Other general features are dynamic and describe the status of a node and its relation to the Voronoi cell it belongs to, including a binary variable denoting whether the node is selected as a facility, the node index, the index of Voronoi cell it belongs to, and the distance from node $i$ to the assigned facility.
If a node is currently selected as a facility, we can exploit additional information from the Voronoi cell it creates. Specifically, we compute three cell-based features for facility $i$: the sum of demand in cell $\sum_{u\in P(V_r)} p_u$, the total cost in cell $\sum_{u\in P(V_r)} p_u d_{f(r)u}$, and the area of polygon that the Voronoi cell forms $s_r$. The facility features are padded with zero for non-facility nodes.
To capture the distances on graphs, we initialize the edge feature between node $i$ and $j$ with the length of the edge, i.e. $\mathbf{e}_{ij}=d_{ij}$.

We employ multi-head graph attention networks for node embedding. Specifically, we adopt the implementation from~\cite{brody2021attentive}, which fixes the static attention problem of the standard GAT~\cite{velivckovic2017graph}. Formally, for each single attention head, 
\begin{align}
    \mathbf{h}_i^{l}&= \mathbf{GAT}(\mathbf{h}_i^{l-1}) \nonumber \\
    &= \mathrm{ReLU}( \alpha_{i, i} \Theta_s \mathbf{h}_i^{l-1}+\sum_{j \in \mathcal{N}(i)} \alpha_{i, j} \Theta_t \mathbf{h}_j^{l-1}), \\
    \alpha_{i, j} &= \frac{ \exp(u_{i, j})}{\sum\limits_{k \in \mathcal{N}(i) \cup\{i\}}  \exp (u_{i,k})}, \\
    u_{i, j} &= \mathbf{a}^{\top} \mathrm { LeakyReLU }\left(\boldsymbol{\Theta}_s \mathbf{h}_i+\boldsymbol{\Theta}_t \mathbf{h}_j+\boldsymbol{\Theta}_e \mathbf{e}_{i, j}\right)
\end{align}
where $\mathcal{N}(i)= \{j \in V|(j, i) \in E\}$ is the neighbors of node $i$, and $\mathbf{a}, \boldsymbol{\Theta}_s, \boldsymbol{\Theta}_t, \boldsymbol{\Theta}_e$ are learnable parameters of GAT. Different attention heads are concatenated. 

\subsubsection{Attention-based Relocation Pair Selection}

The action space for one relocation involves choosing two nodes: the facility to remove $u_1$ and the new facility to insert $u_2$, As the action space is quadratic to the number of nodes $(u_1, u_2) \in F \times  (V\setminus F)$, we break down the target into a two-stage task, i.e. $P(u_1, u_2|\Theta_{\mathcal{A}}) = P(u_1|\Theta_{\mathcal{A}}) P(u_2|\Theta_{\mathcal{A}}, u_1)$. 

After the GAT module, each node $i$ has an embedding vector $\mathbf{h}_i^{L_1}$. 
The following MLP module serves the purpose of information compression and acts as a scorer, which evaluates the priority of removing each facility. Each layer is defined by
\begin{align}
	\mathbf{g}_i^l &= \mathbf{MLP}(\mathbf{g}_i^{l-1}) = \mathrm{ReLU}(\mathrm{Linear}(\mathbf{g}_i^{l-1})).
\end{align}
After $L_2$ layers of MLPs, each node has a scalar of score, which yields the final probability distribution of removing $u_1$:
\begin{align}
	&P(u_1|\Theta_{\mathcal{A}}) = \sigma(logit_1(u_1)), \\
	&logit_1(u_1) = \left\{
	\begin{aligned}
		\mathbf{g}_i^{L_2}& , & u_1 \in F \\
		-\infty & , & u_1 \notin F
	\end{aligned}
	\right.
	,
\end{align} 
where $\sigma$ represents the Softmax function. 

Next we consider choosing the new facility $u_2$ given node embeddings and the removed facility $u_1$. This is implemented by an attention layer
\begin{align}
	&\mathbf{f}_i = \mathbf{Att}(\mathbf{h}_i^{L_1}) = (\mathbf{h}_{i}^{L_1})^{\top} \mathrm{tanh}(\mathrm{Linear}(\mathbf{h}_{u_1}^{L_1})), \\
	&P(u_2|\Theta_{\mathcal{A}}, u_1) = \sigma(logit_2(u_2)), \\
	&logit_2(u_2) = \left\{
	\begin{aligned}
		\mathbf{f}_i & , & u_2 \in V\setminus F \\
		-\infty & , & u_2 \notin V\setminus F
	\end{aligned}
	\right.
	.
\end{align}
Now that we have the probability of choosing $u_1$ and $u_2$, the agent samples according to $P(u_1|\Theta_{\mathcal{A}})$ and $P(u_2|\Theta_{\mathcal{A}}, u_1)$ when taking actions.

\subsubsection{Imitation Learning}
The large action space gives PPO-swap a high-level of autonomy, yet increasing the difficulty for training. 
As introduced in Section~\ref{sec:preliminaries}, interchange algorithms, such as Greedy-swap, exhibit exceptional performance. Greedy-swap can generate high-quality solutions within acceptable running times, particularly for small instances. Therefore, we propose leveraging imitation learning to guide and expedite the training of neural networks. First the Greedy-swap algorithm is run on small instances, and we collect the interaction trajectories between the Greedy-swap agent and the environment, which are referred to as the \textit{expert data}. These data are then used to supervise the training of the actor network. The pre-trained model is then fine-tuned for different training dataset using the PPO algorithm. We adopt the classic \textit{behavioral cloning} algorithm for imitation learning.

\begin{algorithm}[tb]
    \caption{Swap algorithm for $p$-median problem}
    \label{alg:swap_pmp}
    
    \SetKwInOut{Param}{Parameters}
	\Param{iteration number $T$, swap number $S$, swapping model $agent$}
    \KwData{facility number $p$}
    \KwResult{facility set $F$}
    
    $F^* \leftarrow \emptyset$\tcp*{best facility location}
    $cost_{min} \leftarrow \infty$\;
    
    \For{$i \leftarrow 1$ \KwTo $T$}{
        $F \leftarrow$ an initial set of $p$ facilities\;
        $(F_k, J_k), F \leftarrow \FSwap(1, agent, F, S)$\;
        \If{$\mathcal{O}(F) < cost_{min}$}{
            $F^* \leftarrow F$\;
			$cost_{min} \leftarrow \mathcal{O}(F)$\;
		}
    }
    \Return $F^*$\;
\end{algorithm}

\subsection{From Relocation to P-median Problem}
\label{sec:pmp}

Improving algorithms not only solve the relocation problem, but can also be applied to the traditional $p$-median problem, so that an initial feasible solution evolves towards lower value of the objective function. Algorithm~\ref{alg:general_swap} can be readily deployed as a subroutine in an interchange framework for solving the $p$-median problem, as shown in Algorithm~\ref{alg:swap_pmp}. With a hyperparameter $S$ controlling the number of swaps, the function \texttt{SwapRelocate} is called after initialization, and the best solution yielded during $T$ trials is returned as the final solution.
All four variants of Algorithm~\ref{alg:general_swap} are capable of transplantation. Specially, if we use the Greedy-swap agent for \texttt{SwapRelocate}, then we end up with the classical Teitz-Bart algorithm~\cite{teitzHeuristicMethods1968}, a well-studied heuristic~\cite{aryaLocalSearch2001}. This showcases the compatibility of our proposed swap framework and justifies the design. 

An important component from relocation to the $p$-median problem is a proper way of initialization. The most straightforward way of initialization is randomly choose $p$ nodes from the graph. Intuitively, a good initial solution is beneficial to the quality of final solution as it is closer to the global/local minimum. 
Here we propose a density-based initialization, which is inspired by the scaling law introduced in Section~\ref{sec:power_law}. Since the population density $\rho$ and facility density $D$ satisfies $D \propto \rho^{2/3}$, it is sensible to set the probability of choosing each node proportional to $\rho^{2/3}$. The default initialization method used in all our experiments is density initialization.


\section{Experiments}
The proposed methods are implemented and evaluated on two tasks: the facility relocation problem and the $p$-median problem on graphs, with two different datasets for each task, namely grid cities and Gabriel graphs. We also assess the generalizability of our model, which is ignored in previous works~\cite{matisReinforcementLearning2023}.

\subsection{Datasets}
We build two sets of synthetic graph data with different structural features, namely grid cities and Gabriel graphs. The generation process includes the 2D coordinates of nodes, the connectivity between nodes, and the demand at each point. 

\subsubsection{Grid Cities}
\begin{figure}[tb]
	\centering
	\includegraphics[width=.5\textwidth]{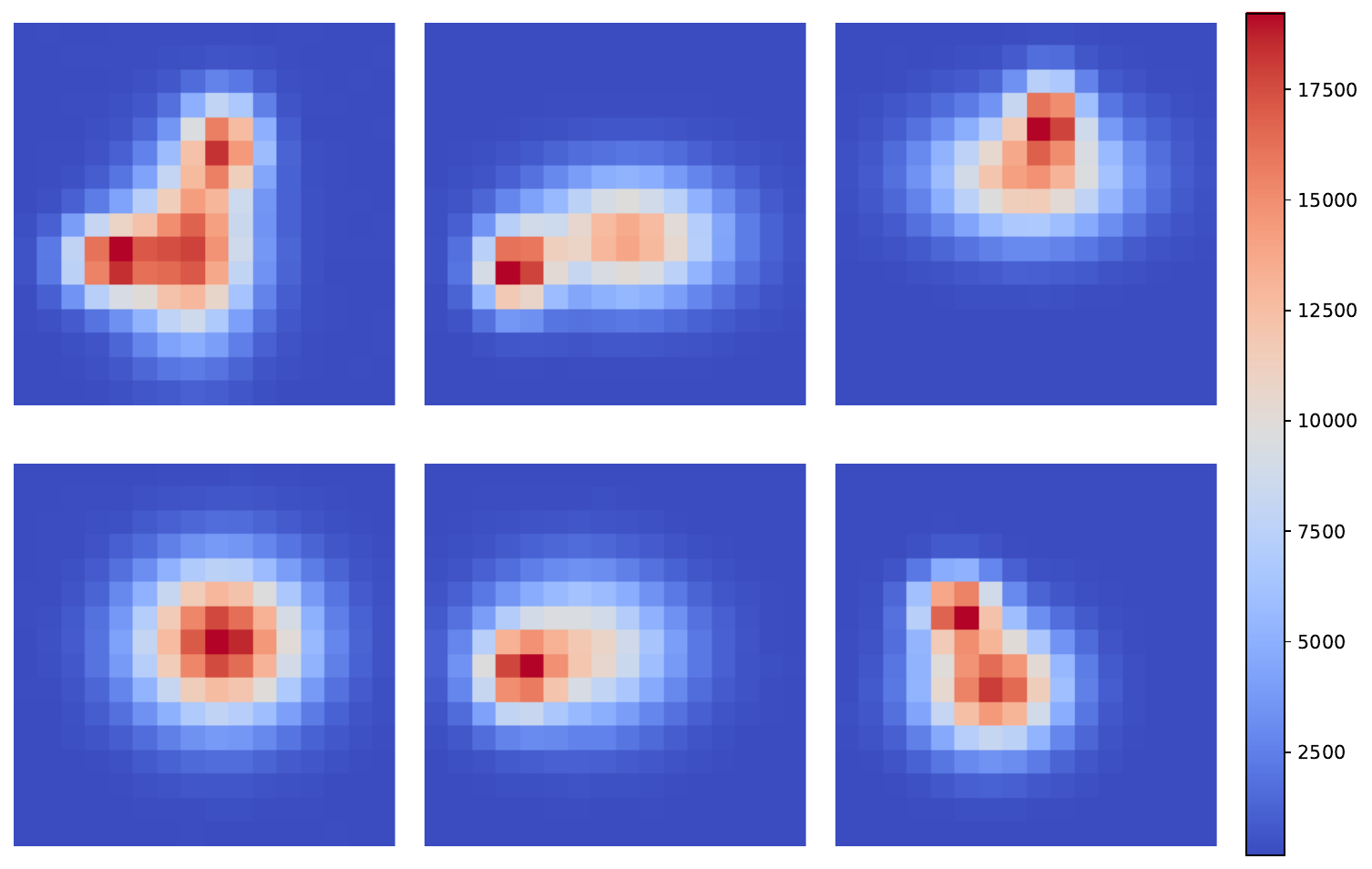}
	\caption{Population distribution of generated grid cities of size $16\times 16$. Red areas indicate more population and higher demand. }
	\label{fig:toy_city}
\end{figure}

Grid cities are intended to simulate the population distribution in road networks with grid patterns. We extend the dataset of synthetic grid cities in~\cite{xuDeconstructingLaws2020} by introducing a multi-center population distribution for more complexity and diversity. For a given city size $w$, the nodes in this city form a $w \times w$ grid, with 8-adjacent nodes connected by an edge, simulating well-structured cities like New York City, Toronto, and Beijing. As for population generation, we randomly generate 1 to 3 central business districts (CBDs), each represented by a bivariate normal distribution in the x-axis and y-axis with random centers and variances. A total population of 500,000 is assigned to the CBDs, and an additional noise of 10\% is uniformly distributed in the grid city. Fig.~\ref{fig:toy_city} demonstrates the population distribution in different examples of grid cities. 

\subsubsection{Gabriel Graphs}
\begin{figure*}[tb] 
    \centering
    \hfil
    \subfloat[]
    {
        \includegraphics[width=0.3\textwidth]{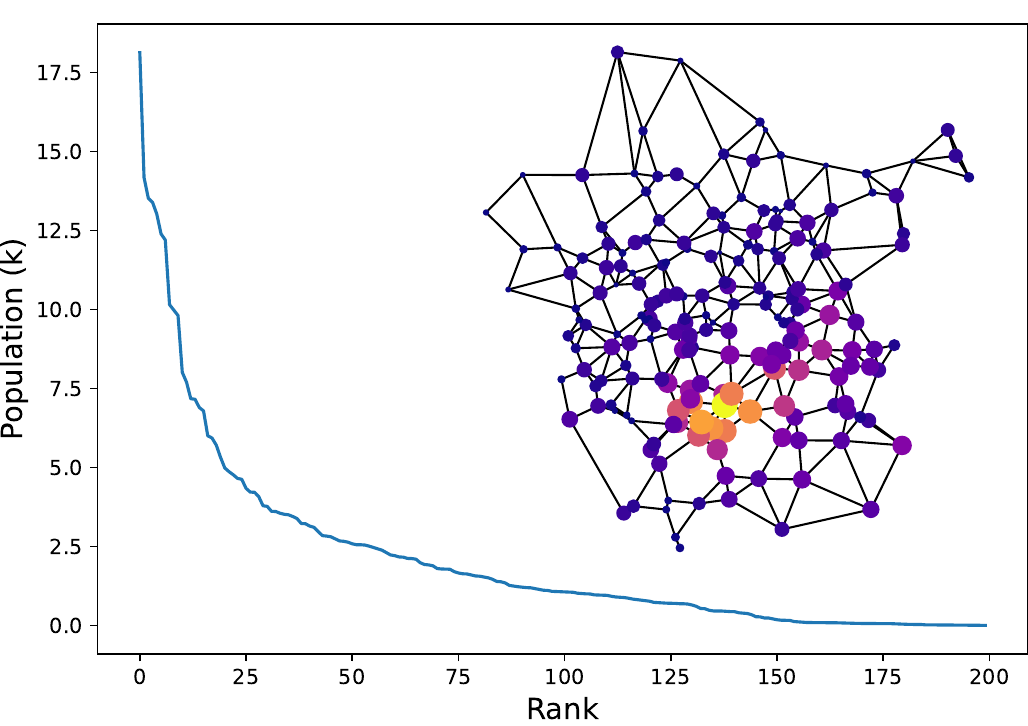}
    }
    \hfil
    \subfloat[]
    {
        \includegraphics[width=0.3\textwidth]{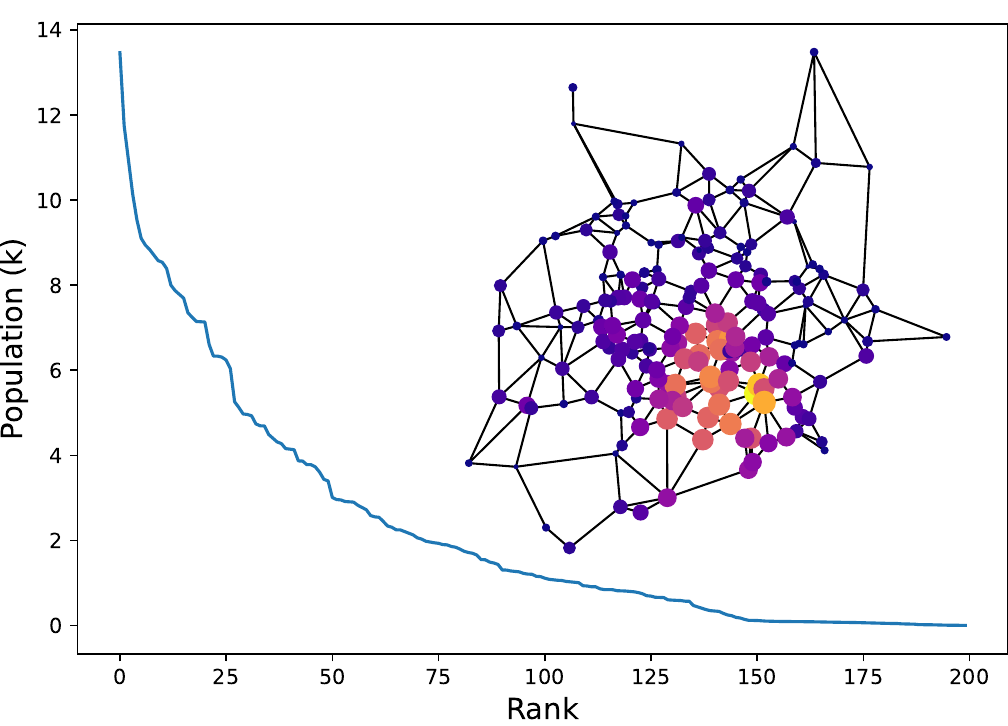}
    }
	\hfil
    \subfloat[]
    {
        \includegraphics[width=0.3\textwidth]{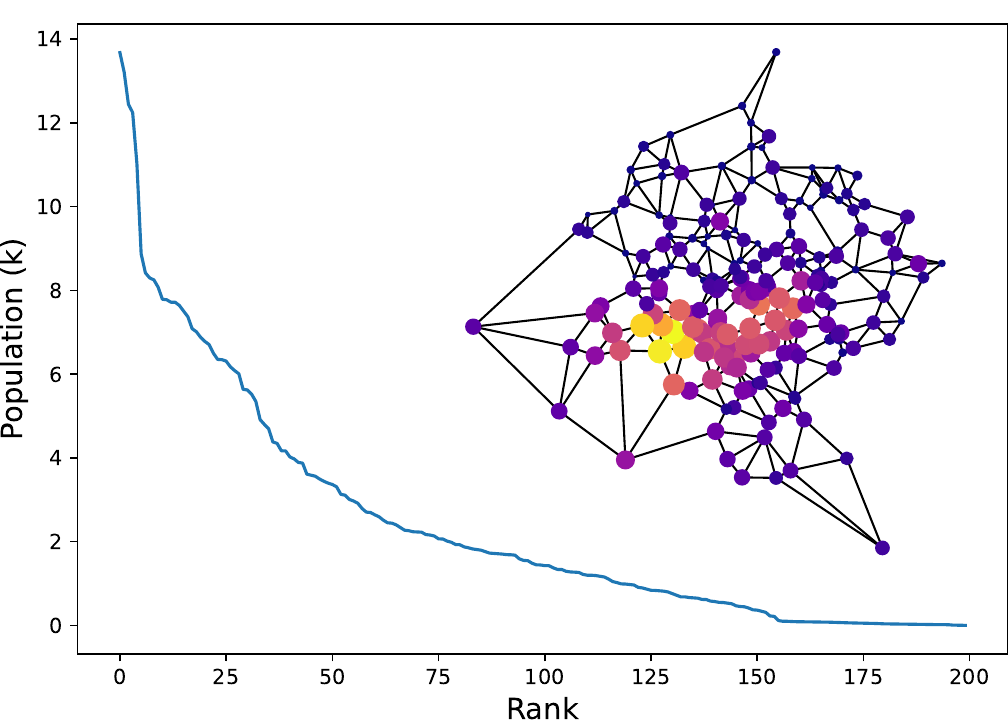}
    }
    \hfil
    \subfloat[]
    {
        \includegraphics[width=0.3\textwidth]{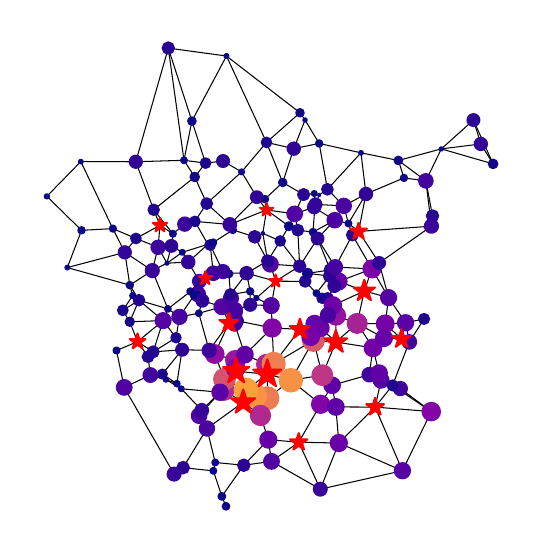}
    }
    \hfil
    \subfloat[]
    {
        \includegraphics[width=0.4\textwidth]{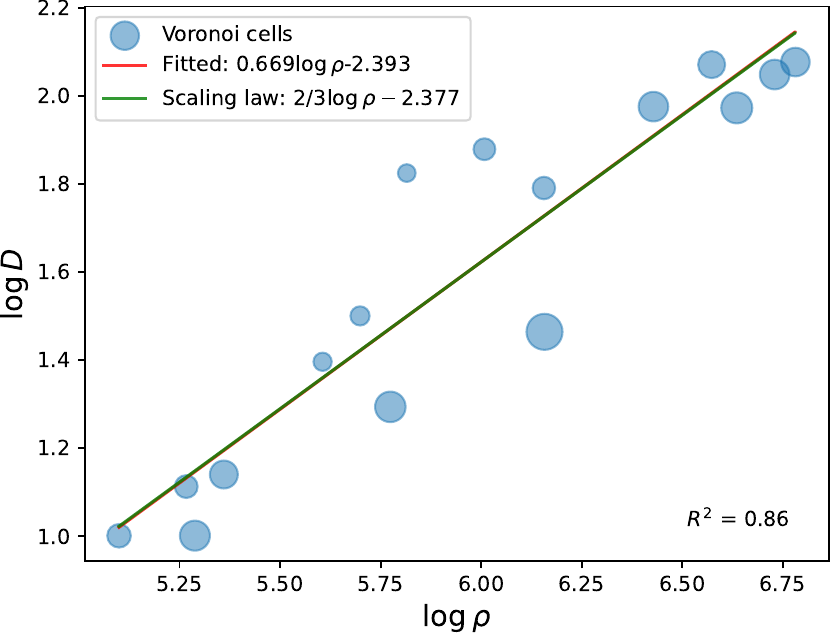}
        \label{fig:gabriel_law}
    }
    \hfil
    \caption{Demonstration of generated Gabriel graph dataset. 
	    \textbf{(a-c)} Example of generated Gabriel graphs with 200 nodes. The node size and color are proportional to the population on nodes. The population-rank plot is obtained by sorting the population of all nodes descendingly. 
	    \textbf{(d)} The stars represent the optimal facility locations on graph (a) when $p=16$. 
	    \textbf{(e)} Verification of the scaling law between population density $\rho$ and facility density $D$ on graph (a) with facilities indicated in (d). We compute $\rho$ and $D$ of each Voronoi cell and illustrate the log-log scatter plot. Linear regression of the data is represented by the red line, highly coinciding with the theoretical scaling law indicated by the green line. }
    \label{fig:gabriel_all}
\end{figure*}

To simulate more irregular network structures, such as road networks in Boston and Shanghai, as well as the communication and power networks, we propose a generation process based on Gabriel graphs~\cite{gabriel1969new}, as it describes the geometric proximity in the Euclidean space. 
Compared to other methods to generate planar graphs, such as the Delaunay triangulation, the node degrees of Gabriel graphs better portray the degree distribution of road networks. 
We first generate $n$ nodes with 2D coordinates following a bivariate normal distribution from $[0,1]^2$ and create edges according to the definition of Gabriel graphs. 
However, nodes tend to form disconnected clusters when $n$ is large. Therefore, we additionally link the edges between each node and its $k$ nearest neighbors within the limit of a predefined degree (a random integer between 3 and 6). The demand (population) on each node is randomly generated with the mean value of eigenvector centrality, a common measure of node influence in the graph. Fig.~\ref{fig:gabriel_all} demonstrates examples of Gabriel graph with 200 nodes.

It is crucial to highlight that the graph generation pipeline we employ is not arbitrary. The scaling law between facility density and population density in Section~\ref{sec:power_law} can be verified on our generated graphs, as shown in Fig.~\ref{fig:gabriel_law}, which substantiates the non-ad hoc nature of our methodology. 
Consequently, we posit that this dataset serves as a robust representation of intricate urban networks and population distribution.

\subsection{Solving Facility Relocation on Graphs}
For the facility relocation problem on graphs with $n$ nodes, we randomly generate a facility set of size $p$ and allow a maximum number of relocation of $\left\lfloor p/2\right\rfloor$. We compare four variants of Algorithm~\ref{alg:general_swap} mentioned in Section~\ref{sec:swap_algs}, including two heuristics \textit{Random-swap} and \textit{Greedy-swap}, as well as our proposed methods \textit{VSCA} and \textit{PPO-swap}. For each graph, the initial facility set is constructed using density initialization in~\ref{sec:pmp}. 

Table~\ref{tab:toy_imp_res} and Table~\ref{tab:gabriel_imp_res} present the performance and running time of solving facility relocation on grid cities and Gabriel graphs, respectively. The improvement ratio $Q$ is defined in (\ref{eq:improvement}), the larger the better. 
VSCA and PPO-swap consistently achieve an improvement ratio over 10\% for all given graphs and initial facility layout, and reach around 20\% on Gabriel graphs of different sizes. For grid cities, the benefits of relocation gained by our proposed methods are close to Greedy-swap, which can be seen as approximation of optimal relocation plans. For VSCA, it strikes a good balance between running time and performance, as it produces high quality solutions with running time lower than Random-swap. This indicates the efficacy of the guidance from Voronoi cells and their regional cost. 
As for PPO-swap, the agent effectively learns to select the relocation pair and reaches even higher improvement ratio among the large action space, proving the learning ability of our proposed model architecture. 
Both VSCA and PPO-swap are able to produce good solutions within reasonable running time (under 6 seconds per instance), while Greedy-swap sacrifices the efficiency and runs extremely slow as the graph becomes larger. 

\begin{table*}[tb]
    \centering
    \caption{Results for Facility Relocation on Grid Cities}
    \label{tab:toy_imp_res}
    \begin{tabular}{lcccccccc}
        \toprule
        \multirow{2}{*}{Methods} & \multicolumn{2}{c}{($n=64, p=6$)} & \multicolumn{2}{c}{($n=64, p=8$)} & \multicolumn{2}{c}{($n=256, p=25$)} & \multicolumn{2}{c}{($n=256, p=39$)} \\
        & $Q$ (\%) & Time (s) & $Q$ (\%) & Time (s) & $Q$ (\%) & Time (s) & $Q$ (\%) & Time (s) \\
        \midrule
		Random-swap & 6.77 & 0.0017 & 1.86 & 0.0022 & 3.65 & 0.0123 & 3.68 & 0.0201 \\
		Greedy-swap & 16.65 & 0.0461 & 13.71 & 0.0789 & 15.64 & 11.6373 & 17.07 & 29.8753 \\
		\textbf{VSCA} & 13.94 & 0.0017 & 11.83 & 0.0015 & 10.35 & 0.0086 & 12.56 & 0.0093 \\
		\textbf{PPO-swap} & 15.62 & 0.1522 & 11.08 & 0.2016 & 13.37 & 0.9937 & 12.72 & 1.7152 \\
        \bottomrule
    \end{tabular}
\end{table*}

\begin{table*}[tb]
    \centering
    \caption{Results for Facility Relocation on Gabriel Graphs}
    \label{tab:gabriel_imp_res}
    \begin{tabular}{lcccccccccc}
        \toprule
        \multirow{2}{*}{Methods} & \multicolumn{2}{c}{($n=100, p=10$)} & \multicolumn{2}{c}{($n=100, p=15$)} & \multicolumn{2}{c}{($n=200, p=20$)} & \multicolumn{2}{c}{($n=200, p=30$)} & \multicolumn{2}{c}{($n=500, p=50$)} \\
        & $Q$ (\%) & Time (s) & $Q$ (\%) & Time (s) & $Q$ (\%) & Time (s) & $Q$ (\%) & Time (s) & $Q$ (\%) & Time (s) \\
        \midrule
		Random-swap & 11.44 & 0.0033 & 9.56 & 0.0043 & 7.81 & 0.0091 & 4.36 & 0.0132 & 4.18 & 0.0533 \\
		Greedy-swap & 27.19 & 0.3225 & 29.92 & 0.6928 & 27.41 & 5.0258 & 30.70 & 10.1836 & 30.78 & 179.9406 \\
		\textbf{VSCA} & 19.11 & 0.0049 & 21.99 & 0.0042 & 18.62 & 0.0134 & 13.76 & 0.0083 & 15.03 & 0.0573 \\
		\textbf{PPO-swap} & 20.81 & 0.3864 & 22.60 & 0.5296 & 19.12 & 0.9503 & 19.96 & 1.4677 & 16.75 & 5.4724 \\
        \bottomrule
    \end{tabular}
\end{table*}

\begin{figure}[tb]
	\centering
	\includegraphics[width=.47\textwidth]{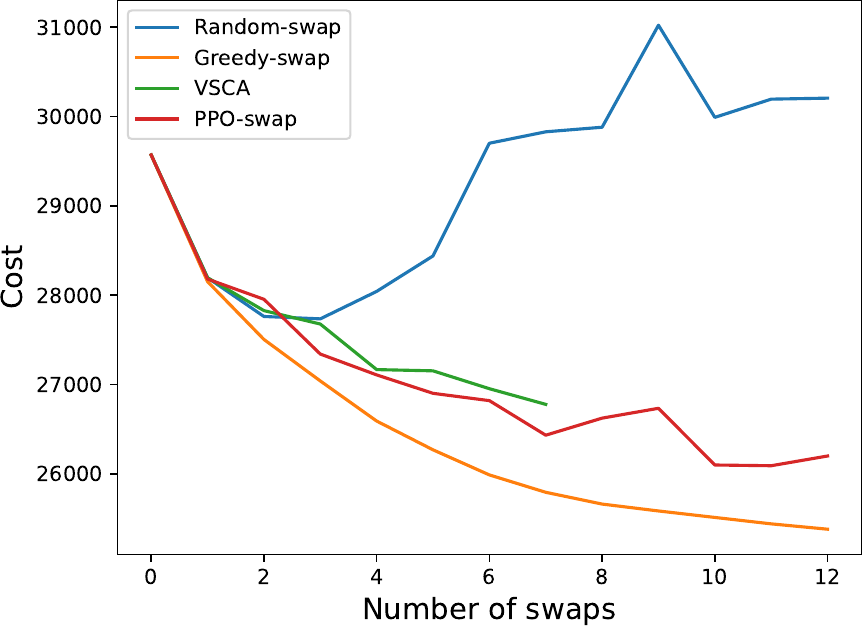}
	\caption{The change of total cost as the relocation processes on a $16\times16$ grid city by different methods. The initial facility set has 25 facilities. }
	\label{fig:relocation_curve}
\end{figure}

Fig.~\ref{fig:relocation_curve} provides a tangible example of how different methods approach the relocation task. It illustrates the relocation process on a $16\times16$ grid city by different methods with 16 initial facilities. 
Random-swap makes decisions blindly and goes in the wrong direction, while VSCA and PPO-swap both stay on the right track. VSCA stops when the cross-region swaps no longer make improvements, but PPO-swap is able to consider long-term benefits and make larger improvements to the end.


\subsection{Solving P-median Problem on Graphs}

As an extended application of the relocation problem, we evaluate the swap-based methods for the facility location problem. 
For solving the $p$-median problem from scratch, we compare with various methods, including the Gurobi optimizer~\cite{gurobi} that finds exact solutions and other approximated heuristics. \textit{Random} serves as a naive baseline by sampling $p$ random nodes from the graph uniformly. \textit{K-means} is an intuitive method based on clustering. It partitions the nodes into $p$ groups, with the nodes closest to cluster centers selected as facilities. \textit{Maranzana} and \textit{Greedy-addition} are two classical heuristics introduced in Section~\ref{sec:pmp_heuristics}, and the rest are swap-based methods, four variants of Algorithm~\ref{alg:swap_pmp}. 

Table~\ref{tab:toy_gap_res} and Table~\ref{tab:gabriel_gap_res} demonstrate the results of solving $p$-median problem on grid cities and Gabriel graphs. The metric is optimality gap, defined as the cost difference between the solution and the optimal solution (produced by Gurobi) divided by the optimal cost. Lower gap is better.
It is interesting to note that K-means is a strong baseline on grid cities and reaches lower optimal gap than the expert heuristic Maranzana. However, its performance deteriorates to a large extent on Gabriel graphs. It is reasonable since the K-means clustering is based on node coordinates and Euclidean distances. When encountering complex graph instances, K-means suffers from the inconsistency between graph distance and straight-line distance. 
The swap-based methods circumvents this problem and outperform K-means on all Gabriel graphs. 
As for VSCA and PPO-swap, they reach comparable results to the Greedy-addition on grid cities with the optimality gap under 9\%, and outperform the Maranzana heuristic in all cases. 
When compared within the swap-based framework, VSCA and PPO-swap speeds up the solving process of Greedy-swap for hundreds of times on large graphs, and are able to make sensible decisions for the swap operations compared to Random-swap. 

\begin{table*}[tb]
    \centering
    \caption{Results for P-Median Problem on Grid Cities}
    \label{tab:toy_gap_res}
    \begin{tabular}{lcccccccc}
        \toprule
        \multirow{2}{*}{Methods} & \multicolumn{2}{c}{($n=64, p=6$)} & \multicolumn{2}{c}{($n=64, p=8$)} & \multicolumn{2}{c}{($n=256, p=25$)} & \multicolumn{2}{c}{($n=256, p=39$)} \\
        & Gap (\%) & Time (s) & Gap (\%) & Time (s) & Gap (\%) & Time (s) & Gap (\%) & Time (s) \\
		\midrule
		Gurobi & 0 & 0.0750 & 0 & 0.0460 & 0 & 9.5089 & 0 & 4.5745 \\
		Random & 33.94 & 0.0009 & 26.56 & 0.0005 & 41.97 & 0.0013 & 41.57 & 0.0012 \\
		K-means & 5.68 & 0.0266 & 7.05 & 0.0263 & 9.99 & 0.0907 & 16.66 & 0.0871 \\
		Maranzana & 5.14 & 0.0081 & 7.56 & 0.0109 & 12.91 & 0.0508 & 18.36 & 0.0551 \\
		Greedy-addition & 0.96 & 0.0173 & 0.99 & 0.0232 & 2.24 & 0.9255 & 2.14 & 1.4612 \\
		\cmidrule{1-9}\morecmidrules\cmidrule{1-9}
		Random-swap & 10.43 & 0.0546 & 8.11 & 0.0638 & 13.69 & 0.4610 & 13.74 & 0.6391 \\
		Greedy-swap & 0.03 & 0.4411 & 0.11 & 0.6986 & 0.64 & 84.8473 & 0.60 & 160.6521 \\
		\textbf{VSCA} & 0.81 & 0.0890 & 1.32 & 0.0904 & 4.40 & 0.6021 & 4.55 & 0.7044 \\
		\textbf{PPO-swap} & 1.26 & 0.3888 & 0.98 & 0.4063 & 7.92 & 1.9959 & 8.82 & 1.8604 \\		
		
        \bottomrule
    \end{tabular} 
\end{table*}

\begin{table*}[tb]
    \centering
    \caption{Results for P-Median Problem on Gabriel Graphs}
    \label{tab:gabriel_gap_res}
    \begin{tabular}{lcccccccccc}
        \toprule
        \multirow{2}{*}{Methods} & \multicolumn{2}{c}{($n=100, p=10$)} & \multicolumn{2}{c}{($n=100, p=15$)} & \multicolumn{2}{c}{($n=200, p=20$)} & \multicolumn{2}{c}{($n=200, p=30$)} & \multicolumn{2}{c}{($n=500, p=50$)} \\
        & Gap (\%) & Time (s) & Gap (\%) & Time (s) & Gap (\%) & Time (s) & Gap (\%) & Time (s) & Gap (\%) & Time (s) \\
		\midrule
		Gurobi & 0 & 0.2359 & 0 & 0.2196 & 0 & 2.3407 & 0 & 1.5201 & 0 & 10.9262 \\
		Random & 52.82 & 0.0005 & 56.97 & 0.0005 & 63.22 & 0.0015 & 69.04 & 0.0017 & 74.97 & 0.0025 \\
		K-means & 24.79 & 0.0576 & 46.93 & 0.0493 & 36.46 & 0.1648 & 52.57 & 0.1796 & 49.17 & 0.0765 \\
		Maranzana & 12.46 & 0.0146 & 25.83 & 0.0293 & 20.84 & 0.0375 & 34.18 & 0.0558 & 28.38 & 0.2086 \\
		Greedy-addition & 3.82 & 0.0642 & 3.31 & 0.1160 & 4.35 & 0.8197 & 4.74 & 1.2464 & 6.20 & 6.2811 \\
		\cmidrule{1-11}\morecmidrules\cmidrule{1-11}
		Random-swap & 23.32 & 0.1466 & 30.90 & 0.2248 & 31.40 & 0.2413 & 34.94 & 0.3214 & 39.63 & 1.4421 \\
		Greedy-swap & 0.05 & 2.8026 & 0.32 & 7.2531 & 0.35 & 38.2104 & 0.93 & 69.3944 & 0.41 & 1456.7650 \\
		\textbf{VSCA} & 6.08 & 0.2173 & 13.74 & 0.2570 & 12.16 & 0.4135 & 17.41 & 0.4529 & 19.88 & 2.3522 \\
		\textbf{PPO-swap} & 5.94 & 1.3086 & 11.64 & 1.3993 & 12.37 & 1.9420 & 15.39 & 2.0321 & 21.90 & 7.3882 \\

        \bottomrule
    \end{tabular}
\end{table*}

\subsection{Model Generalizability}
\begin{figure}[tb]
	\centering
	\includegraphics[width=0.47\textwidth]{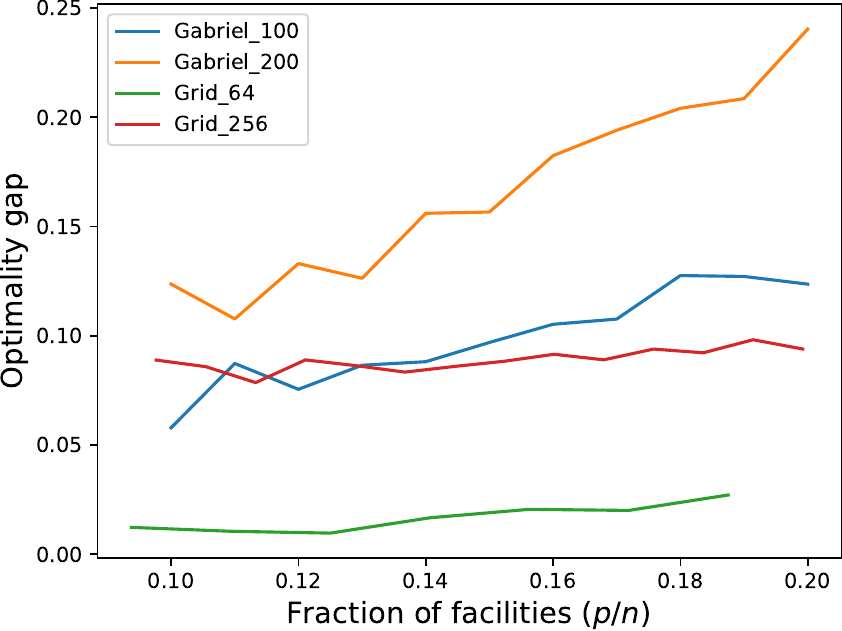}
	\caption{The optimality gap of PPO-swap solving the $p$-median problem for different number of facilities. The tested model is trained on dataset of a fixed $p = 0.1n$.}
	\label{fig:generalization}
\end{figure}
We further conduct the experiments on generalizability of our reinforcement learning model by evaluating its performance as the number of facilities changes. Results are presented in~\ref{fig:generalization}. For each dataset, the model is trained with data only containing a fixed number of $p=0.1n$, that is, $p$ equals to 10\% of the nodes. However, PPO-swap is able to generalize to different scales of problems without seeing this kind of problems. On grid cities, the model performs equally well for harder instances as $p$ increases to $0.2n$.

\subsection{Implementation Details}

\begin{table}[tb]
  \centering
  \caption{Hyperparameters for Training PPO-swap}
  \label{tab:hyperparameters}
  \begin{tabular}{|c|c|}
    \hline
    \textbf{Hyperparameters} & \textbf{Value} \\
	\hline
    PPO gamma & 0.9 \\
    \hline
    GAE lambda & 0.95 \\
    \hline
    Batch size & 64 \\
	\hline
	Steps per epoch & 1024 \\
	\hline
	Optimization iterations & 4 \\
	\hline
	Clip ratio & 0.1 \\
	\hline
	Clip decay & 0.998 \\
	\hline
	Entropy loss weight & 0.01 \\
	\hline
	Critic loss weight & 0.5 \\
	\hline
	Gradient clipping & 0.5 \\
	\hline
	Number of epochs & 500 \\
	\hline
  \end{tabular}
\end{table}

For Grid cities dataset, we use two different sizes of cities: $8\times 8$ and $16\times 16$ denoted by Grid\_64 and Grid\_256. For Gabriel graphs dataset, three different scales of graphs are used, and we denote them as Gabriel\_100, Gabriel\_200, and Gabriel\_500.

The training process of PPO-swap contains two steps: imitation learning and PPO algorithm for reinforcement training. The expert data of imitation learning come from the trajectory of Greedy-swap on Grid\_256 and Gabriel\_100, which are then used to train two base models: Grid\_IL and Gabriel\_IL. The models are then fine-tuned with PPO algorithm on each dataset with the Adam optimizer, with a learning rate of 0.0001 and exponential decay rate of 0.99. 
The $\mathbf{GAT}$ module and $\mathbf{MLP}$ modules both have three layers, with hidden size of 64 and output dimensions of 128. The $\mathbf{GAT}$ uses 4 heads attention.
The environment for RL training use fixed $p=0.1n$ for each dataset. Other hyperparameters are listed in Table~\ref{tab:hyperparameters}.
The test dataset are generated from the same distribution, each with 10 instances. For all nondeterministic methods, we set the iteration number $T=5$ and the best solutions are reported.

\section{Conclusion}
In conclusion, our work provides a robust solution to the challenges posed by facility location problems on graphs. We introduce a versatile swap-based framework addressing both the $p$-median problem and facility relocation on graphs, achieving a commendable balance between solution quality and running time. Outperforming handcrafted heuristics, our approach stands out on intricate graph datasets. Additionally, our physics-inspired initialization strategy enhances optimization efficiency, and our graph generation process aligns with urban distribution patterns. 
Together, these advancements provide a holistic and effective solution to facility location problems on graphs, advancing the understanding and resolution of facility location problems on graphs. 
Our methods can be potentially applied to urban dataset of real cities, addressing practical challenges of facility location problems on graphs. 

\bibliographystyle{IEEEtran}
\bibliography{main}

\begin{thebibliography}{10}
\providecommand{\url}[1]{#1}
\csname url@samestyle\endcsname
\providecommand{\newblock}{\relax}
\providecommand{\bibinfo}[2]{#2}
\providecommand{\BIBentrySTDinterwordspacing}{\spaceskip=0pt\relax}
\providecommand{\BIBentryALTinterwordstretchfactor}{4}
\providecommand{\BIBentryALTinterwordspacing}{\spaceskip=\fontdimen2\font plus
\BIBentryALTinterwordstretchfactor\fontdimen3\font minus \fontdimen4\font\relax}
\providecommand{\BIBforeignlanguage}[2]{{%
\expandafter\ifx\csname l@#1\endcsname\relax
\typeout{** WARNING: IEEEtran.bst: No hyphenation pattern has been}%
\typeout{** loaded for the language `#1'. Using the pattern for}%
\typeout{** the default language instead.}%
\else
\language=\csname l@#1\endcsname
\fi
#2}}
\providecommand{\BIBdecl}{\relax}
\BIBdecl

\bibitem{celikturkogluComparativeSurvey2020}
D.~Celik~Turkoglu and M.~Erol~Genevois, ``A comparative survey of service facility location problems,'' \emph{Annals of Operations Research}, vol. 292, no.~1, pp. 399--468, 2020.

\bibitem{farahaniFacilityLocation2009}
R.~Z. Farahani and M.~Hekmatfar, \emph{Facility Location: Concepts, Models, Algorithms and Case Studies}.\hskip 1em plus 0.5em minus 0.4em\relax Springer Science \& Business Media, 2009.

\bibitem{GAVRANOVIC201415}
\BIBentryALTinterwordspacing
H.~Gavranović, A.~Barut, G.~Ertek, O.~B. Yüzbaşıoğlu, O.~Pekpostalcı, and Önder Tombuş, ``Optimizing the electric charge station network of eŞarj,'' \emph{Procedia Computer Science}, vol.~31, pp. 15--21, 2014, 2nd International Conference on Information Technology and Quantitative Management, ITQM 2014. [Online]. Available: \url{https://www.sciencedirect.com/science/article/pii/S1877050914004177}
\BIBentrySTDinterwordspacing

\bibitem{ndiayeApplicationPMedian2012}
F.~Ndiaye, B.~M. Ndiaye, and I.~Ly, ``Application of the p-median problem in school allocation,'' \emph{American Journal of Operations Research}, vol.~02, no.~02, pp. 253--259, 2012.

\bibitem{CINTRANO2020113684}
\BIBentryALTinterwordspacing
C.~Cintrano, F.~Chicano, and E.~Alba, ``Using metaheuristics for the location of bicycle stations,'' \emph{Expert Systems with Applications}, vol. 161, p. 113684, 2020. [Online]. Available: \url{https://www.sciencedirect.com/science/article/pii/S095741742030508X}
\BIBentrySTDinterwordspacing

\bibitem{bengioMachineLearning2020}
\BIBentryALTinterwordspacing
Y.~Bengio, A.~Lodi, and A.~Prouvost, ``Machine learning for combinatorial optimization: A methodological tour d’horizon,'' \emph{European Journal of Operational Research}, vol. 290, no.~2, pp. 405--421, 2021. [Online]. Available: \url{https://www.sciencedirect.com/science/article/pii/S0377221720306895}
\BIBentrySTDinterwordspacing

\bibitem{wangSolvingUncapacitated2022}
C.~Wang, C.~Han, T.~Guo, and M.~Ding, ``Solving uncapacitated p-median problem with reinforcement learning assisted by graph attention networks,'' \emph{Applied Intelligence}, 2022.

\bibitem{matisReinforcementLearning2023}
D.~Matis and P.~Tar{\'a}bek, ``Reinforcement learning for weighted p-median problem,'' in \emph{2023 International Conference on Information and Digital Technologies (IDT)}, 2023, pp. 293--298.

\bibitem{luoFacilityRelocation2023}
H.~Luo, Z.~Bao, J.~S. Culpepper, M.~Li, and Y.~Zhao, ``Facility relocation search for good: When facility exposure meets user convenience,'' in \emph{Proceedings of the ACM Web Conference 2023}, ser. WWW '23.\hskip 1em plus 0.5em minus 0.4em\relax New York, NY, USA: Association for Computing Machinery, 2023, pp. 3937--3947.

\bibitem{tsiotasTopologyUrban2017}
D.~Tsiotas and S.~Polyzos, ``The topology of urban road networks and its role to urban mobility,'' \emph{Transportation Research Procedia}, vol.~24, pp. 482--490, 2017.

\bibitem{daiLearningCombinatorial2018}
E.~Khalil, H.~Dai, Y.~Zhang, B.~Dilkina, and L.~Song, ``Learning combinatorial optimization algorithms over graphs,'' in \emph{Advances in Neural Information Processing Systems}, I.~Guyon, U.~V. Luxburg, S.~Bengio, H.~Wallach, R.~Fergus, S.~Vishwanathan, and R.~Garnett, Eds., vol.~30.\hskip 1em plus 0.5em minus 0.4em\relax Curran Associates, Inc., 2017.

\bibitem{kool2019attention}
\BIBentryALTinterwordspacing
W.~Kool, H.~van Hoof, and M.~Welling, ``Attention, learn to solve routing problems!'' in \emph{7th International Conference on Learning Representations, {ICLR} 2019, New Orleans, LA, USA, May 6-9, 2019}.\hskip 1em plus 0.5em minus 0.4em\relax OpenReview.net, 2019. [Online]. Available: \url{https://openreview.net/forum?id=ByxBFsRqYm}
\BIBentrySTDinterwordspacing

\bibitem{cappartCombinatorialOptimization2022}
\BIBentryALTinterwordspacing
Q.~Cappart, D.~Ch{\'e}telat, E.~Khalil, A.~Lodi, C.~Morris, and P.~Veli{\v c}kovi{\'c}, ``Combinatorial optimization and reasoning with graph neural networks,'' \emph{Journal of Machine Learning Research}, vol.~24, no. 130, pp. 1--61, 2023. [Online]. Available: \url{http://jmlr.org/papers/v24/21-0449.html}
\BIBentrySTDinterwordspacing

\bibitem{chenLearningPerform2019}
X.~Chen and Y.~Tian, ``Learning to perform local rewriting for combinatorial optimization,'' in \emph{Advances in Neural Information Processing Systems}, H.~Wallach, H.~Larochelle, A.~Beygelzimer, F.~d\textquotesingle Alch\'{e}-Buc, E.~Fox, and R.~Garnett, Eds., vol.~32.\hskip 1em plus 0.5em minus 0.4em\relax Curran Associates, Inc., 2019.

\bibitem{luLearningbasedIterative2019}
H.~Lu, X.~Zhang, and S.~Yang, ``A learning-based iterative method for solving vehicle routing problems,'' in \emph{International Conference on Learning Representations}, 2019.

\bibitem{wuLearningImprovement2022}
Y.~Wu, W.~Song, Z.~Cao, J.~Zhang, and A.~Lim, ``Learning improvement heuristics for solving routing problems,'' \emph{IEEE Transactions on Neural Networks and Learning Systems}, vol.~33, no.~9, pp. 5057--5069, 2022.

\bibitem{falknerLearningControl2023}
J.~K. Falkner, D.~Thyssens, A.~Bdeir, and L.~{Schmidt-Thieme}, ``Learning to control local search for combinatorial optimization,'' in \emph{Machine Learning and Knowledge Discovery in Databases}, M.-R. Amini, S.~Canu, A.~Fischer, T.~Guns, P.~Kralj~Novak, and G.~Tsoumakas, Eds.\hskip 1em plus 0.5em minus 0.4em\relax Cham: Springer Nature Switzerland, 2023, vol. 13717, pp. 361--376.

\bibitem{hansen1997variable}
P.~Hansen and N.~Mladenovi{\'c}, ``Variable neighborhood search for the p-median,'' \emph{Location Science}, vol.~5, no.~4, pp. 207--226, 1997.

\bibitem{alp2003efficient}
O.~Alp, E.~Erkut, and Z.~Drezner, ``An efficient genetic algorithm for the p-median problem,'' \emph{Annals of Operations research}, vol. 122, pp. 21--42, 2003.

\bibitem{rolland1997efficient}
E.~Rolland, D.~A. Schilling, and J.~R. Current, ``An efficient tabu search procedure for the p-median problem,'' \emph{European Journal of Operational Research}, vol.~96, no.~2, pp. 329--342, 1997.

\bibitem{cornuejols1977exceptional}
G.~Cornuejols, M.~L. Fisher, and G.~L. Nemhauser, ``Exceptional paper—location of bank accounts to optimize float: An analytic study of exact and approximate algorithms,'' \emph{Management science}, vol.~23, no.~8, pp. 789--810, 1977.

\bibitem{gurobi}
\BIBentryALTinterwordspacing
{Gurobi Optimization, LLC}, ``{Gurobi Optimizer Reference Manual},'' 2023. [Online]. Available: \url{https://www.gurobi.com}
\BIBentrySTDinterwordspacing

\bibitem{kuehn1963heuristic}
A.~A. Kuehn and M.~J. Hamburger, ``A heuristic program for locating warehouses,'' \emph{Management science}, vol.~9, no.~4, pp. 643--666, 1963.

\bibitem{maranzana1964location}
F.~Maranzana, ``On the location of supply points to minimize transport costs,'' \emph{Journal of the Operational Research Society}, vol.~15, no.~3, pp. 261--270, 1964.

\bibitem{goodchild1983location}
M.~F. Goodchild and V.~T. Noronha, \emph{Location-allocation for small computers}.\hskip 1em plus 0.5em minus 0.4em\relax Department of Geography, University of Iowa, 1983, no.~8.

\bibitem{reeseSolutionMethods2006}
J.~Reese, ``Solution methods for the p-median problem: An annotated bibliography,'' \emph{Networks}, vol.~48, no.~3, pp. 125--142, 2006.

\bibitem{williams1992simple}
R.~J. Williams, ``Simple statistical gradient-following algorithms for connectionist reinforcement learning,'' \emph{Machine learning}, vol.~8, pp. 229--256, 1992.

\bibitem{shaohuawangDeepMCLPSolving2023}
{Shaohua Wang}, {Haojian Liang}, Y.~Zhong, {Xueyan Zhang}, and C.~Su, ``Deepmclp: Solving the mclp with deep reinforcement learning for urban facility location analytics,'' \emph{Spatial Data Science Symposium 2023}, 2023.

\bibitem{dijkstraNoteTwo1959}
E.~W. Dijkstra, ``A note on two problems in connexion with graphs,'' \emph{Numerische Mathematik}, vol.~1, no.~1, pp. 269--271, 1959.

\bibitem{kariv1979algorithmic}
O.~Kariv and S.~L. Hakimi, ``An algorithmic approach to network location problems. i: The p-centers,'' \emph{SIAM journal on applied mathematics}, vol.~37, no.~3, pp. 513--538, 1979.

\bibitem{gwalaniEvaluationHeuristics2021}
H.~Gwalani, C.~Tiwari, and A.~R. Mikler, ``Evaluation of heuristics for the p-median problem: Scale and spatial demand distribution,'' \emph{Computers, Environment and Urban Systems}, vol.~88, p. 101656, 2021.

\bibitem{umScalingLaws2009}
J.~Um, S.-W. Son, S.-I. Lee, H.~Jeong, and B.~J. Kim, ``Scaling laws between population and facility densities,'' \emph{Proceedings of the National Academy of Sciences}, vol. 106, no.~34, pp. 14\,236--14\,240, 2009.

\bibitem{gastnerOptimalDesign2006}
M.~T. Gastner and M.~E.~J. Newman, ``Optimal design of spatial distribution networks,'' \emph{Physical Review E}, vol.~74, no.~1, p. 016117, 2006.

\bibitem{schulman2017proximal}
J.~Schulman, F.~Wolski, P.~Dhariwal, A.~Radford, and O.~Klimov, ``Proximal policy optimization algorithms,'' \emph{arXiv preprint arXiv:1707.06347}, 2017.

\bibitem{huang202237}
S.~Huang, R.~F.~J. Dossa, A.~Raffin, A.~Kanervisto, and W.~Wang, ``The 37 implementation details of proximal policy optimization,'' \emph{The ICLR Blog Track 2023}, 2022.

\bibitem{brody2021attentive}
\BIBentryALTinterwordspacing
S.~Brody, U.~Alon, and E.~Yahav, ``How attentive are graph attention networks?'' in \emph{The Tenth International Conference on Learning Representations, {ICLR} 2022, Virtual Event, April 25-29, 2022}.\hskip 1em plus 0.5em minus 0.4em\relax OpenReview.net, 2022. [Online]. Available: \url{https://openreview.net/forum?id=F72ximsx7C1}
\BIBentrySTDinterwordspacing

\bibitem{velivckovic2017graph}
\BIBentryALTinterwordspacing
P.~Veli{\v{c}}kovi{\'c}, G.~Cucurull, A.~Casanova, A.~Romero, P.~Lio, and Y.~Bengio, ``Graph attention networks,'' in \emph{6th International Conference on Learning Representations, {ICLR} 2018, Vancouver, BC, Canada, April 30 - May 3, 2018, Conference Track Proceedings}.\hskip 1em plus 0.5em minus 0.4em\relax OpenReview.net, 2018. [Online]. Available: \url{https://openreview.net/forum?id=rJXMpikCZ}
\BIBentrySTDinterwordspacing

\bibitem{teitzHeuristicMethods1968}
M.~B. Teitz and P.~Bart, ``Heuristic methods for estimating the generalized vertex median of a weighted graph,'' \emph{Operations Research}, vol.~16, no.~5, pp. 955--961, 1968.

\bibitem{aryaLocalSearch2001}
V.~Arya, N.~Garg, R.~Khandekar, A.~Meyerson, K.~Munagala, and V.~Pandit, ``Local search heuristic for k-median and facility location problems,'' in \emph{Proceedings of the Thirty-Third Annual ACM Symposium on Theory of Computing}.\hskip 1em plus 0.5em minus 0.4em\relax Hersonissos Greece: ACM, 2001, pp. 21--29.

\bibitem{xuDeconstructingLaws2020}
Y.~Xu, L.~E. Olmos, S.~Abbar, and M.~C. Gonz{\'a}lez, ``Deconstructing laws of accessibility and facility distribution in cities,'' \emph{Science Advances}, vol.~6, no.~37, p. eabb4112, 2020.

\bibitem{gabriel1969new}
K.~R. Gabriel and R.~R. Sokal, ``A new statistical approach to geographic variation analysis,'' \emph{Systematic zoology}, vol.~18, no.~3, pp. 259--278, 1969.

\end{thebibliography}

\end{document}